\documentclass[journal]{IEEEtran}
\usepackage{amsmath,amsfonts}
\usepackage{algorithm}
\usepackage{algpseudocode}
\usepackage{array}
\usepackage[caption=false,font=normalsize,labelfont=sf,textfont=sf]{subfig}
\usepackage{textcomp}
\usepackage{stfloats}
\usepackage{url}
\usepackage{verbatim}
\usepackage{booktabs}
\usepackage{xcolor}
\usepackage{makecell}
\usepackage[hidelinks]{hyperref}
\usepackage{graphicx}
\usepackage{multirow}
\usepackage{cite}

\newtheorem{definition}{Definition}
\newtheorem{property}{Property}
\newcounter{example}
\begin{document}

\title{Multi-Agent Agentic Graph Learning via \\ Structural Signatures}

\author{
    Liang Qu$^{1}$~\IEEEmembership{Member,~IEEE},
    Jianxin Li$^{1\ast}$~\IEEEmembership{Senior Member,~IEEE},
    Hua~Wang$^{2}$~\IEEEmembership{Fellow,~IEEE}
    \thanks{$^{1}$Edith Cowan University, Perth, Australia. Liang Qu: leoqu94@outlook.com.}%
    \thanks{$^{2}$The University of Queensland, Brisbane, Australia.}%
    \thanks{$^{\ast}$Corresponding author: Jianxin Li (jianxin.li@ecu.edu.au).}%
}

% The paper headers
\markboth{}%
{Qu \MakeLowercase{\textit{et al.}}: Multi-Agent Agentic Graph Learning via Structural Signatures}

% \IEEEpubid{0000--0000/00\$00.00~\copyright~2021 IEEE}
% Remember, if you use this you must call \IEEEpubidadjcol in the second
% column for its text to clear the IEEEpubid mark.

\maketitle

\begin{abstract}
Agentic graph learning (AGL) has recently achieved promising results on graph reasoning tasks, where an agent powered by a large language model (LLM) sequentially samples the graph as evidence to support its final prediction.
Existing methods either employ a single agent or orchestrate multiple role-based agents to reason and learn over the entire graph, but both essentially rely on a shared reasoning policy across different graph regions, which can be suboptimal for graphs with heterogeneous structural and semantic patterns.
Inspired by the progress of multi-agent collaboration on complex reasoning tasks, a natural remedy is to let multiple agents own different memory and collaborate; however, applying this paradigm to graphs directly faces two challenges.
First, existing AGL methods typically verbalize graph structures into natural-language descriptions for LLM agents, making the reasoning process sensitive to the ordering of structural information and thereby breaking the permutation-invariant nature of graphs.
Second, incorporating increasingly large sampled neighborhoods leads to rapidly growing contexts, which not only increases inference cost but also makes important structural evidence vulnerable to the lost-in-the-middle problem.
To address these challenges, this paper introduces a multi-agent agentic graph learning (i.e., \textsc{MAAGL}) framework. \textsc{MAAGL} partitions the graph into communities and assigns an independent agent to each community for region-specific specialization. Unlike existing methods that verbalize all sampled evidence into text, \textsc{MAAGL} represents structural and semantic evidence separately. Structural evidence is summarized by a dynamically updated \emph{structural signature} that is permutation-invariant and fixed in size, while semantic evidence is filtered to the top-$k$ nodes ranked by relevance. Based on historical trajectories with similar signatures, agents estimate their confidence and trigger debate-style collaboration when needed. \textsc{MAAGL} further learns reusable experience from historical trajectories to guide subsequent reasoning. Extensive experiments on four benchmark datasets show that \textsc{MAAGL} outperforms state-of-the-art AGL methods under both in-domain and zero-shot transfer settings while reducing token overhead.
\end{abstract}


\section{Introduction}

Text-attributed graphs (TAGs) are widely used to model relational data in many real-world applications \cite{xia2026graph,yang2025graphagent,sun2026agentgl}. A TAG consists of structural information and semantic information. The structural information is represented by nodes and edges, where nodes denote entities and edges describe the relations between them. The semantic information is the text attached to each node. For example, in social networks \cite{tan2019deep}, nodes represent users and edges represent their social connections, while user profiles provide the text of the nodes. In citation networks \cite{gao2024heterogeneous}, nodes represent papers and edges represent citation relations between papers, while titles, abstracts, and keywords provide the text of the papers. As a result, the core objective of learning on TAGs is to learn from the structural information and the semantic information jointly.

The traditional method to learn from both kinds of information is graph neural networks (GNNs) \cite{wu2020comprehensive}. A GNN starts from the semantic information of each node, encoded as a feature vector, and passes it along the edges. At each layer, a node aggregates the representations of its neighbors and updates its own. Thus, after several layers, a node representation carries the semantics of its multi-hop neighborhood. Early models such as GCN \cite{kipf2016semi}, GraphSAGE \cite{hamilton2017inductive}, and GAT \cite{velivckovic2017graph} mainly differ in how the neighbors are selected and how their information are aggregated. Recently, graph transformers such as Graphormer \cite{ying2021transformers} and GraphGPS \cite{rampavsek2022recipe} are proposed to add attention over the whole graph for capturing long-range dependencies.
Despite their success, these methods share a fundamental limitation, i.e., the same neighborhood selection and aggregation rule is applied uniformly across all nodes. But, in fact, different nodes may require different neighborhood contexts for downstream tasks. 

\begin{figure*}[t]
    \centering
    \begin{minipage}[c]{0.45\textwidth}
        \centering
        \includegraphics[width=\linewidth]{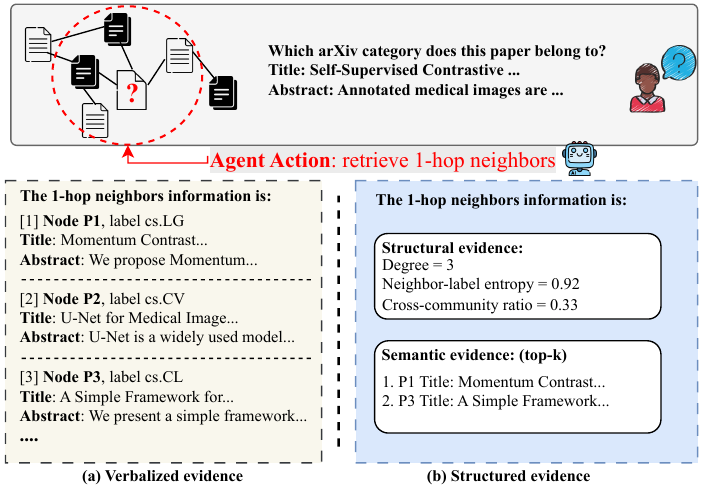}
    \end{minipage}\hfill
    \begin{minipage}[c]{0.49\textwidth}
        \centering
        \includegraphics[width=\linewidth]{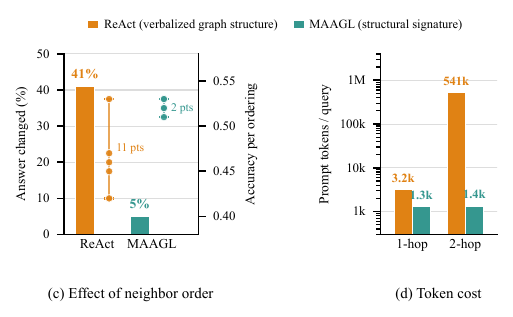}
    \end{minipage}
    \caption{(a) Agents reasoning with verbalized graph evidence. (b) Agents reasoning with the proposed structural evidence. (c) Neighbor-order sensitivity of a ReAct-based agent \cite{yao2022react} on OGB-Arxiv \cite{hu2020open} across five independent runs, where the retrieved neighbors are presented in different random orders. (d) Prompt tokens per query on the same nodes.}
    \label{fig:ordersens}
\end{figure*}

Recently, agentic graph learning (AGL) \cite{bei2025graphs} has emerged as a new paradigm that leverages the planning and reasoning capabilities of LLMs together with graph-specific tools to adaptively identify and collect relevant structural and semantic evidence for downstream graph tasks.
Existing AGL methods can be roughly divided into two categories. Single-agent methods \cite{sun2026agentgl,guo2025reagan} employ a single LLM agent to iteratively retrieve graph evidence and reason toward the final prediction. Orchestration-based methods \cite{yang2025graphagent,wang2026graphcogent} instead decompose this reasoning process across multiple role-specific agents and coordinate them through a predefined workflow, with different agents responsible for different stages of graph reasoning. Despite this architectural difference, both paradigms essentially rely on a shared reasoning policy over the entire graph.
However, as the graph becomes larger and more heterogeneous, mixing diverse structural and semantic patterns into a shared policy can be suboptimal.

To address the above issue, a natural solution is to introduce multi-agent collaborative reasoning (e.g., multi-agent debate \cite{du2023improving}), which has shown strong effectiveness on complex reasoning tasks by leveraging diverse perspectives from agents with independent memories and reasoning experiences \cite{du2023improving,hao2026self}. However, directly applying such methods to graph reasoning tasks is non-trivial and faces two key challenges. (1) Existing multi-agent collaboration is mainly developed for language-based reasoning tasks, where both the problem context and reasoning process are naturally expressed in text, making textual exchange between agents straightforward. In graph reasoning, however, part of the evidence lies in the graph structure, which must first be verbalized into a textual sequence for agent communication, as illustrated in Figure~\ref{fig:ordersens} (a). This verbalization imposes an arbitrary order on inherently unordered graph neighborhoods, breaking permutation invariance and making the reasoning sensitive to neighbor ordering. As evidenced by our preliminary results in Figure~\ref{fig:ordersens} (c), different orderings of the same neighbors can lead to different predictions. (2) Verbalizing graph structures also causes the context to grow rapidly as the neighborhood expands. In TAGs, each retrieved node brings not only structural information but also its associated textual attributes, all of which need to be included in the textual context. Due to the combinatorial expansion of multi-hop neighborhoods, the resulting context can therefore grow rapidly with each additional hop, leading to substantial token costs, as evidenced by our preliminary results in Figure~\ref{fig:ordersens}(d).

In light of the above challenges, this work proposes a multi-agent agentic graph learning \textsc{MAAGL} framework. Specifically, to enable agent specialization, \textsc{MAAGL} first partitions the graph into communities and assigns one independent agent to each community. Each agent then performs independent reasoning within its assigned community to build region-specific reasoning experience.
Unlike existing methods that verbalize all sampled graph evidence into text, \textsc{MAAGL} represents the evidence returned by each reasoning action in two parts: structural evidence and semantic evidence. For structural evidence, we summarize the observed neighborhood using a \emph{structural signature} composed of a small set of graph statistics (e.g., node degree and neighborhood label entropy). The signature is dynamically updated over the sampled node set as new evidence is collected, while remaining permutation-invariant and fixed in size. For semantic evidence, \textsc{MAAGL} retains the textual attributes of only the top-$k$ retrieved nodes ranked by their semantic relevance to the target node. In this way, \textsc{MAAGL} preserves useful structural and semantic evidence while avoiding arbitrary neighbor ordering and excessive token costs.
During subsequent reasoning, each agent estimates its confidence using the success rate of past trajectories with similar structural signatures. If the confidence is below a threshold, the same score is computed for the other agents, and the top-$K$ agents are selected for debate-style collaborative reasoning. Finally, \textsc{MAAGL} learns from historical reasoning trajectories and stores the learned experiences in agent memory to guide subsequent reasoning. The contributions of this work can be summarized as follows:
\begin{itemize}
    \item We identify a fundamental mismatch in applying existing multi-agent collaborative reasoning to graph reasoning tasks. Directly verbalizing sampled graph evidence for agent reasoning can break permutation invariance and incur high token costs.
    \item We propose \textsc{MAAGL}, a multi-agent agentic graph learning framework. \textsc{MAAGL} separates sampled graph evidence into structural and semantic information, representing the former with a dynamically updated \emph{structural signature} and filtering the latter by semantic relevance.
    \item We conduct extensive experiments on four benchmark datasets, where \textsc{MAAGL} outperforms AGL methods under both in-domain and zero-shot transfer settings while reducing token overhead.
\end{itemize}

The rest of this paper is organized as follows. Section 2 reviews related work. We formally define the problem of \textsc{MAAGL} in Section 3 and present its framework with technical details in Section 4. Section 5 demonstrates the experimental evaluation and Section 6 concludes the paper.

\section{Related Work}

\subsection{GNN-based Graph Learning}
Graph neural networks (GNNs) \cite{wu2020comprehensive} are the dominant approach for learning from both structural and semantic information on graphs. A GNN starts from the semantic information of each node, encoded as a feature vector, and propagates it along graph edges through message passing, so that each node gradually aggregates information from its local structure. Different GNN architectures mainly differ in how neighbors are selected and how their information is aggregated. GCN \cite{kipf2016semi} performs normalized neighborhood aggregation, GraphSAGE \cite{hamilton2017inductive} samples neighbors and applies a learnable aggregator, GAT \cite{velivckovic2017graph} assigns attention weights to different neighbors, and JK-Net \cite{xu2018representation} combines representations from multiple layers to capture neighborhoods at different ranges. Later work extends message passing to heterogeneous graphs with different node and edge types \cite{wang2019heterogeneous,fu2020magnn}, while graph transformers \cite{yun2019graph,rampavsek2022recipe} introduce global attention to capture long-range dependencies. 
Another line of work improves their training with graph augmentation. DropEdge \cite{rong2019dropedge} randomly removes edges during training, which reduces over-fitting and over-smoothing in deep GNNs. GraphCL \cite{you2020graph} perturbs the graph into several contrastive views and learns representations that agree across the views. NodeAug \cite{wang2020nodeaug} and local augmentation \cite{liu2022local} enrich the surroundings of each node, the former by changing its nearby attributes and edges under consistency training, the latter by generating extra neighbor features conditioned on the node.

However, GNN-based graph learning methods generally rely on neighborhood selection and aggregation rules determined by the model architecture and applied uniformly across nodes, which can be suboptimal when different nodes require different graph contexts for downstream tasks.

\subsection{Agent-based Graph Learning}
Large language model (LLM)-powered agents have recently emerged as a promising paradigm for solving complex tasks by integrating planning, reasoning, and tool use within an autonomous loop \cite{yehudai2025survey,yao2022react}. This paradigm has also been extended to graph learning, giving rise to agentic graph learning (AGL) \cite{bei2025graphs}. Existing AGL methods mainly follow two lines. Single-agent methods use one LLM agent to interact with the graph and iteratively collect evidence for reasoning. ReaGAN \cite{guo2025reagan} equips an agent with graph sampling tools for retrieving structural and semantic evidence such as neighbor labels, while Graph-CoT \cite{jin2024graph} lets an agent iteratively invoke graph functions and reason over the returned information. AgentGL \cite{sun2026agentgl} further organizes graph reasoning into a thought-action-observation loop \cite{yao2022react} and optimizes the reasoning policy from collected trajectories, and GraphReAct \cite{yu2026graphreact} performs multi-step reasoning and acting for graph inference. Orchestration-based methods instead coordinate multiple role-specific agents through a predefined workflow. GraphAgent \cite{yang2025graphagent} assigns different agents to planning, graph retrieval, and prediction, GraphTeam \cite{li2024graphteam} coordinates role-specialized agents to collaboratively solve graph analysis tasks, GraphCogent \cite{wang2026graphcogent} decomposes complex graph understanding across multiple collaborating agents, while GraphMaster \cite{du2025graphmaster} adopts a similar division of labor for graph data synthesis. 

Despite their architectural differences, both single-agent and orchestration-based methods essentially rely on a shared reasoning policy over the entire graph. Although the policy can adaptively collect evidence for individual instances, different graph regions may exhibit distinct structural and semantic patterns. As the graph becomes larger and more heterogeneous, mixing these diverse patterns into a shared policy can therefore be suboptimal.

\subsection{Multi-agent Collaborative Reasoning}
Multi-agent collaborative reasoning has shown promising performance on complex reasoning tasks by combining diverse perspectives from agents with independent memories and reasoning experiences \cite{du2023improving,hao2026self}. Existing methods mainly differ in how agents collaborate. Debate-style methods \cite{du2023improving} involve multiple agents in the same task and iteratively refine their decisions by exchanging reasoning results: each agent reads the answers and rationales of the others, revises its own over several rounds, and the final decision is reached by consensus or voting. Routing-based methods select only a subset of agents according to their expertise for each task; AgentRouter \cite{zhang2025agentrouter} embeds the incoming task and retrieves the agents whose recorded expertise is semantically closest, and follow-up work studies how to select the best set of collaborators under a cost budget \cite{wang2025optimal}. Other methods enable agents to share past experience through a common memory without direct interaction \cite{gao2024memory}. An agent writes what it has learned into a shared pool, from which other agents retrieve when they meet similar tasks. These methods are mainly developed for language-based reasoning tasks, where both the task information and reasoning process can be naturally represented and exchanged in text.

Directly applying such collaboration to graph reasoning, however, is non-trivial. Graph reasoning additionally relies on structural evidence, which must be verbalized into text before it can be exchanged between agents. This verbalization imposes an arbitrary order on inherently unordered graph neighborhoods, breaking permutation invariance and making the reasoning sensitive to neighbor ordering. Moreover, as multi-hop neighborhoods expand, verbalizing the sampled nodes together with their textual attributes can quickly increase the context length, leading to high token costs. 

\section{Problem Formulation}

\textbf{Text-attributed Graph (TAG):} We define a TAG~\cite{yang2025graphagent,sun2026agentgl} as $G = (V, E, \mathcal{X}, \mathcal{Y})$, where $V$ denotes the set of nodes and $E \subseteq V \times V$ denotes the set of edges. Each node $v \in V$ is associated with a textual attribute (e.g., the title and abstract of a paper node in academic citation graphs) $\mathbf{x}_v \in \mathcal{X}$, where $\mathcal{X}$ denotes the textual attribute space. A subset of nodes $V_L \subseteq V$ is labeled, and each labeled node $v \in V_L$ is associated with a class label $y_v \in \mathcal{Y}$, where $\mathcal{Y}$ denotes the label space.

\textbf{Multi-Agent Agentic Graph Learning:}
Given a TAG $G$ and a target node $v$, multi-agent agentic graph learning employs a set of $M$ LLM-powered agents $\mathcal{A}=\{A_1,\ldots,A_M\}$, where each agent $A_i$ maintains its own reasoning policy $\pi_i$ and private memory $\mathcal{M}_i$.
We formulate the reasoning of each agent as a sequential decision process. At step $t$, agent $A_i$ observes the evidence collected so far, denoted by $s_v^{t}$, and selects an action $a_v^{t}=\pi_i(s_v^{t},\mathcal{M}_i)$ from $\mathcal{A}_{\mathrm{evi}} \cup \{a_{\mathrm{pred}}\}$, where $\mathcal{A}_{\mathrm{evi}}$ is a set of evidence-collection actions that sample structural and semantic evidence from $G$, and $a_{\mathrm{pred}}$ terminates the process with a prediction $\hat{y}_v$.
Agents may further collaborate on the same target node, and every reasoning process is recorded as a trajectory in the memory of the corresponding agent.
Given the labeled node set $V_L$, the goal is to enrich the agent memories $\{\mathcal{M}_i\}_{i=1}^{M}$ by reasoning on $V_L$, such that the agents, individually or collaboratively, correctly predict the labels of unlabeled nodes without updating any parameter of the underlying LLM.

\section{The MAAGL framework}
\begin{figure*}
    \centering
    \includegraphics[width=1\textwidth]{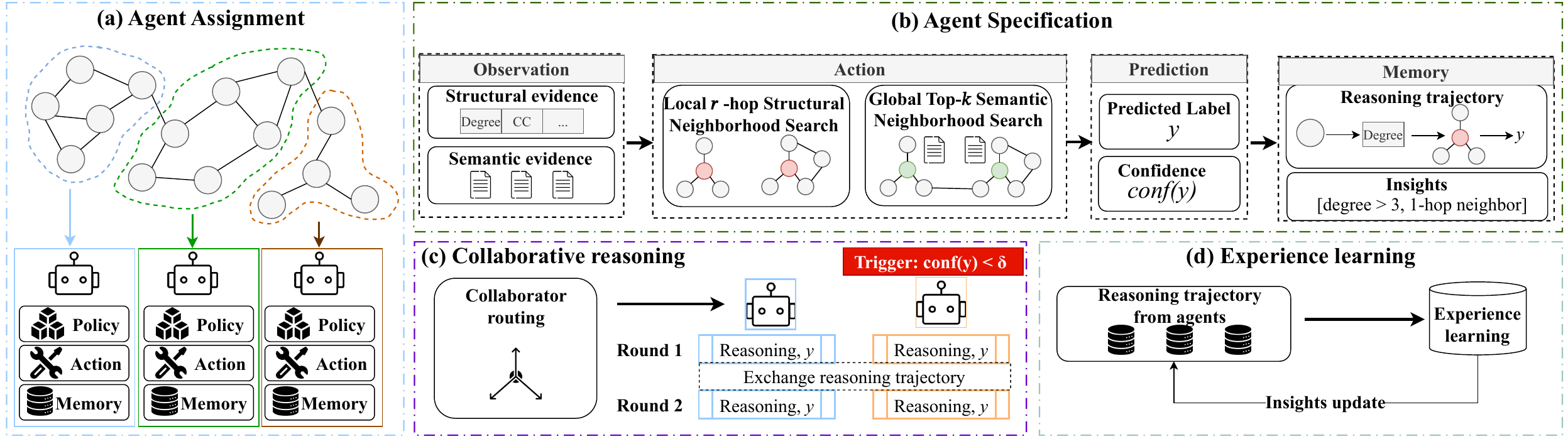}
    \caption{The overall architecture of MAAGL.}
    \label{fig:motivation}
\end{figure*}

As shown in Figure~\ref{fig:motivation}, \textsc{MAAGL} consists of four stages: (a) Agent Assignment (Section~\ref{sec:stage1}), where the graph is partitioned and one agent is assigned to each region; (b) Agent Specialization (Section~\ref{sec:stage2}), where each agent independently builds region-specific reasoning experience; (c) Collaborative Reasoning (Section~\ref{sec:stage3}), where low-confidence cases trigger collaboration among selected agents; and (d) Experience Learning (Section~\ref{sec:stage4}), where historical trajectories are used to learn reusable reasoning experience.

\subsection{Agent Assignment}
\label{sec:stage1}

Existing AGL methods, including single-agent methods \cite{sun2026agentgl,guo2025reagan} and orchestration-based methods \cite{yang2025graphagent,wang2026graphcogent}, typically use a shared reasoning policy and memory across the entire graph. As a result, reasoning experience collected from different graph regions is mixed together, even though these regions may exhibit different structural and semantic patterns. To enable agent specialization, we instead maintain multiple independent agents, each with its own policy and memory, and assign each agent to a specific graph region.

A straightforward approach is to randomly split the nodes among agents. However, such a split ignores graph connectivity and may place closely connected nodes into different regions. We therefore use community detection to group densely connected nodes and assign each community to one agent. Formally,
\begin{equation}
    \{c_1, c_2, \dots, c_M\} = \mathrm{Partition}(G),
    \label{eq:partition}
\end{equation}
where $\mathrm{Partition}(\cdot)$ partitions the node set into $M$ disjoint communities, and agent $A_i$ is assigned to community $c_i$. In this work, we adopt the Leiden algorithm \cite{traag2019louvain} and further study the effect of different partitioning methods in Section \ref{sec:rq3} .

\subsection{Agent Specialization}
\label{sec:stage2}

\subsubsection{Observation}

After agent assignment, each agent first reasons independently within its assigned community to develop region-specific expertise. We formulate this reasoning process as a sequential decision process, where the agent repeatedly takes graph sampling actions and uses the returned evidence to determine its subsequent actions.

A key issue is how the sampled graph evidence is represented to the LLM agent. Existing AGL methods typically verbalize all returned evidence into natural language. For example, after retrieving the 1-hop neighbors of a target node, the node identities, labels, and textual attributes of the sampled neighbors are listed one by one in the prompt. As more graph evidence is sampled, this verbalized context grows accordingly. Example~\ref{ex:verbalized} illustrates such a representation.

\medskip
\noindent\fbox{\parbox{0.96\columnwidth}{\small
\refstepcounter{example}\label{ex:verbalized}%
\textbf{\textit{Example \theexample\ (verbalized evidence).}}\\[5pt]
\hspace*{1.5em}``The 1-hop neighbors information is:\\
\hspace*{1.5em}\phantom{``}[1] Node $u_1$, label $y_{u_1}$, title and abstract ...\\
\hspace*{1.5em}\phantom{``}[2] Node $u_2$, label $y_{u_2}$, title and abstract ...\\
\hspace*{1.5em}\phantom{``}$\dots$\\
\hspace*{1.5em}\phantom{``}[200] Node $u_{200}$, label $y_{u_{200}}$, title and abstract ...''
}}
\medskip

This representation is problematic for graph reasoning. First, graph neighborhoods are inherently unordered, whereas verbalization necessarily places the sampled neighbors into a sequence. The resulting arbitrary order can therefore affect the LLM's reasoning. Second, multi-hop neighborhoods can expand rapidly, and verbalizing every sampled node together with its textual attribute introduces substantial token overhead.

To address this issue, unlike existing methods that uniformly verbalize all sampled graph evidence, \textsc{MAAGL} separates the returned evidence into \emph{structural evidence} and \emph{semantic evidence} and represents them differently. Structural evidence is summarized by a compact \emph{structural signature}, while semantic evidence is filtered according to its relevance to the target node. We first introduce the structural signature.

\begin{definition}[Structural Signature]
\label{def:signature}
The structural signature of a node $v$ at reasoning step $t$ is a vector of $d$ structural statistics,
\begin{equation}
    z_v^{t} =
    [g_1(v,O_v^{t}), g_2(v,O_v^{t}), \dots, g_d(v,O_v^{t})],
    \label{eq:signature}
\end{equation}
where $O_v^{t}$ denotes the set of sampled nodes observed up to step $t$, and each $g_j(\cdot)$ is a deterministic graph statistic.
\end{definition} 

In this work, we use six statistics and divide them into static and dynamic dimensions. The four static dimensions describe the structural position of $v$ and remain unchanged during reasoning. \emph{Degree} measures its local connectivity. \emph{PageRank} measures its global importance under random walks. The \emph{clustering coefficient} measures the connectivity among its 1-hop neighbors. The \emph{cross-community fraction} measures the fraction of 1-hop neighbors that belong to communities different from that of $v$ in Eq.~\eqref{eq:partition}. The remaining two dimensions are updated as the agent samples new evidence. \emph{Label entropy} measures the diversity of labels among the sampled labeled nodes, while the \emph{majority-label fraction} measures the proportion of the most frequent label. Both are computed over the labeled nodes in the sampled node set $O_v^{t}$ of Definition~\ref{def:signature}. Whenever a graph action returns new labeled nodes, the dynamic dimensions are recomputed while the static dimensions keep their initial values,
\begin{equation}
z_v^{t}[j] =
\begin{cases}
g_j(v,O_v^{t}) & \text{if } g_j \text{ is dynamic},\\
z_v^{0}[j] & \text{otherwise}.
\end{cases}
\label{eq:sigupdate}
\end{equation}
Therefore, the signature evolves with the sampled evidence while remaining fixed in dimensionality.

For interaction with the LLM, the numeric signature is rendered into a short textual form that reports the name and value of each statistic,
\begin{equation}
\mathrm{text}(z_v^{t})
=
\big(
\mathrm{name}(g_j),
g_j(v,O_v^{t})
\big)_{j=1}^{d}.
\label{eq:textform}
\end{equation}
In addition to the signature, the observed label counts over the labeled nodes in $O_v^{t}$, from which the two dynamic dimensions are computed, are also reported to the agent.

For semantic evidence, \textsc{MAAGL} does not retain the textual attributes of all sampled nodes. Instead, the sampled nodes are ranked by their semantic relevance to the target node, measured by the cosine similarity between their text embeddings. Only the top-$k$ textual attributes are retained. Unlike the arbitrary ordering of graph neighbors, this order is meaningful because it directly reflects semantic relevance. Accordingly, the observation of agent $A_i$ for target node $v \in c_i$ at step $t$ is represented as
\begin{equation}
s_v^{t}
=
\big(
v,\,
\mathbf{x}_v,\,
\mathrm{text}(z_v^{t}),\,
\mathcal{X}_v^{t,k}
\big),
\label{eq:state}
\end{equation}
where $\mathbf{x}_v$ is the textual attribute of the target node and $\mathcal{X}_v^{t,k}$ denotes the textual attributes of the top-$k$ semantically relevant nodes among the evidence sampled up to step $t$. Example~\ref{ex:maaglstate} illustrates the resulting representation.

\medskip
\noindent\fbox{\parbox{0.96\columnwidth}{\small
\refstepcounter{example}\label{ex:maaglstate}%
\textbf{\textit{Example \theexample\ (MAAGL evidence representation).}}\\[5pt]
\hspace*{1.5em}\textbf{Structural evidence:}\\
\hspace*{2.5em}Degree $=3$\\
\hspace*{2.5em}Neighbor-label entropy $=0.92$\\
\hspace*{2.5em} Observed label counts: Label\_5: 1, Label\_9: 1, Label\_23: 1 \\
\hspace*{2.5em}Cross-community fraction $=0.33$\\[4pt]
\hspace*{1.5em}\textbf{Semantic evidence (top-$k$):}\\
\hspace*{2.5em}1. Node $u_1$: title and abstract ...\\
\hspace*{2.5em}2. Node $u_3$: title and abstract ...

}}
\medskip

The structural signature avoids imposing an arbitrary order on the sampled neighborhood. Since every dimension is computed from the sampled node set rather than its enumeration, it satisfies permutation invariance.

\begin{property}[Permutation Invariance]
\label{prop:perm}
For any node $v$ and sampled node set $O_v^{t}$, the structural signature $z_v^{t}$ is invariant to any permutation of the nodes in $O_v^{t}$.
\end{property}

\begin{IEEEproof}
Let $O_v^{t}=\{u_1,\ldots,u_m\}$ be the sampled node set at step $t$, and let $\sigma$ be any permutation of its elements. For the dynamic dimensions, define the label proportion of class $y\in\mathcal{Y}$ as $p_y(O_v^{t})=|\{u\in O_v^{t}:y_u=y\}|/|O_v^{t}|$. Since permutation does not change the label counts, $p_y(O_v^{t})=p_y(\sigma(O_v^{t}))$ for every class $y$. Hence, both the label entropy $H(O_v^{t})=-\sum_{y\in\mathcal{Y}}p_y(O_v^{t})\log p_y(O_v^{t})$ and the majority-label fraction $F(O_v^{t})=\max_{y\in\mathcal{Y}}p_y(O_v^{t})$ are invariant to node ordering. For the static dimensions, degree depends only on $|\mathcal{N}_1(v)|$, clustering coefficient on the edges among $\mathcal{N}_1(v)$, cross-community fraction on the number of neighbors outside the community of $v$, and PageRank on the graph topology. None of these quantities depends on how the neighbors are enumerated. Therefore, $g_j(v,O_v^{t})=g_j(v,\sigma(O_v^{t}))$ for every $j=1,\ldots,d$, so that $z_v^{t}$ takes the same value for $O_v^{t}$ and $\sigma(O_v^{t})$. Thus, the structural signature is permutation-invariant.
\end{IEEEproof}

At the same time, the dimensionality of $z_v^{t}$ remains fixed regardless of how many nodes have been sampled. Together with top-$k$ filtering of semantic evidence, this representation avoids arbitrary neighbor ordering while preventing the observation from growing directly with the sampled neighborhood size.

\subsubsection{Action}

Given the current observation, the agent selects an evidence-collection action to acquire additional information from the graph. Consistent with the two types of evidence defined above, we consider structural evidence obtained from local graph neighborhoods and semantic evidence obtained from text similarity. Accordingly, the evidence-collection action space is
\begin{equation}
\mathcal{A}_{\mathrm{evi}}
=
\underbrace{\{a_{\mathrm{1\text{-}hop}},\,a_{\mathrm{2\text{-}hop}}\}}_{\text{structural search}}
\cup
\underbrace{\{a_{\mathrm{sem}}\}}_{\text{semantic search}} .
\label{eq:actionspace}
\end{equation}
Each action returns a set of sampled nodes together with their available labels and textual attributes. As described in the previous subsection, the returned evidence is not directly verbalized. Instead, its structural information is incorporated into the structural signature, while its semantic information is filtered by relevance before being presented to the agent.

\begin{definition}[Local $r$-hop Structural Neighborhood Search]
\label{def:rhop}
Given a target node $v$ and a radius $r\in\{1,2\}$, the action $a_{r\text{-}\mathrm{hop}}$ samples the $r$-hop neighborhood
\begin{equation}
\mathcal{N}_r(v)
=
\{u\in V \mid d_G(u,v)=r\},
\label{eq:rhop}
\end{equation}
where $d_G(\cdot,\cdot)$ denotes the shortest-path distance on $G$. The sampled nodes provide structural evidence for updating the signature $z_v^t$. Their textual attributes are further ranked by semantic relevance to the target node, and only the top-$k$ are retained as semantic evidence in the observation.
\end{definition}

\begin{definition}[Global Top-$k$ Semantic Neighborhood Search]
\label{def:sem}
Given a target node $v$, the action $a_{\mathrm{sem}}$ retrieves the $k$ nodes whose textual attributes are most semantically similar to $\mathbf{x}_v$:
\begin{equation}
\mathcal{N}_{\mathrm{sem}}^{k}(v)
=
\operatorname{TopK}_{u\in V\setminus\{v\}}
\cos\!\left(
\mathrm{Enc}(\mathbf{x}_v),
\mathrm{Enc}(\mathbf{x}_u)
\right),
\label{eq:semevidence}
\end{equation}
where $\mathrm{Enc}(\cdot)$ denotes a text encoder. The textual attributes of these nodes form the semantic evidence, while their available labels are also incorporated into the sampled node set used to update the dynamic dimensions of the structural signature.
\end{definition}

\subsubsection{Prediction}
\label{def:pred}

With the observation and evidence-collection actions defined above, each agent independently decides whether to collect more evidence or terminate the reasoning process with a prediction. At step $t$, agent $A_i$ selects an action according to its policy,
\begin{equation}
a_v^{t} = \pi_i(s_v^{t}, \mathcal{M}_i),
\label{eq:policy}
\end{equation}
where $s_v^{t}$ is the current observation and $\mathcal{M}_i$ is the private memory of agent $A_i$. The action is selected from $\mathcal{A}_{\mathrm{evi}} \cup \{a_{\mathrm{pred}}\}$, where $\mathcal{A}_{\mathrm{evi}}$ contains the evidence-collection actions defined above.

If an evidence-collection action is selected, the returned nodes are added to the sampled node set. The structural signature is then updated by Eq.~\eqref{eq:sigupdate}, while the semantic evidence is re-ranked and filtered to retain the top-$k$ relevant textual attributes. The next observation becomes
\begin{equation}
s_v^{t+1}
=
\big(
v,\,
\mathbf{x}_v,\,
\mathrm{text}(z_v^{t+1}),\,
\mathcal{X}_v^{t+1,k}
\big).
\label{eq:stateupdate}
\end{equation}
The agent then continues reasoning based on the updated evidence. When $a_{\mathrm{pred}}$ is selected, the reasoning process terminates and agent $A_i$ produces
\begin{equation}
\hat{y}_v = A_i\big(s_v^{t},\mathcal{M}_i\big).
\label{eq:predict}
\end{equation}
We allow at most $T$ reasoning steps. If the budget is exhausted before $a_{\mathrm{pred}}$ is selected, the final step is used for prediction. We set $T=3$ in our experiments. Fig.~\ref{fig:prompt_reasoning} shows the prompt template used at each reasoning step, which presents the structural signature, the filtered semantic evidence, the search history, and the available experiences to the agent.

\begin{figure}[t]
\centering
\fbox{\parbox{0.96\columnwidth}{\footnotesize\ttfamily\raggedright
You are an agent responsible for one community of a \{graph kind\}. Your task is to classify the target \{item\} into one of the candidate categories.\\[4pt]
Target \{item\}: "\{node text\}"\\[4pt]
Structural signature of the target node, computed by the system and updated as evidence is sampled:\\
\{for each dimension $j$: name($g_j$) = $g_j(v, O_v^t)$ (description); an updated dimension also shows its previous value\}\\
- observed label counts: \{label: count, ...\}\\[4pt]
Semantic evidence, the sampled nodes most relevant to the target (best match first):\\
\{top-$k$ sampled texts, each with its label\}\\[4pt]
Search history:\\
\{one line per earlier action, with the newly observed label counts\}\\[4pt]
Experiences available to you:\\
\{for each experience: IF \{metric\} > \{threshold\} THEN take the \{action\} action\}\\[4pt]
Candidate categories: [\{label list\}]\\[4pt]
Choose exactly one action:\\
1-hop: sample the 1-hop neighbors of the target node.\\
2-hop: sample the 2-hop neighborhood; newly observed labels update the signature.\\
semantic: retrieve the most semantically similar labeled nodes in the whole graph.\\
predict: output the final label based on the current context.\\[4pt]
Respond strictly in JSON: \{"thought": "...", "action": "1-hop|2-hop|semantic|predict", "label": "one candidate category, required only when action is predict"\}
}}
\caption{Prompt template for one reasoning step of an agent. Curly braces denote placeholders filled by the system at runtime.}
\label{fig:prompt_reasoning}
\end{figure}

\noindent\textbf{Adaptive Search Termination.}
Different instances may require different amounts of graph evidence. Some nodes can be resolved from their own attributes or a small local neighborhood, whereas others require additional structural or semantic evidence. Requiring every instance to perform the same number of searches can therefore introduce redundant evidence collection. Besides increasing token cost, unnecessary searches may bring irrelevant nodes into the reasoning context and interfere with the final prediction.

Inspired by the Search-Constrained Thinking strategy of AgentGL \cite{sun2026agentgl}, we encourage each agent to stop searching once the collected evidence is sufficient. Rather than learning this behavior through reinforcement learning, we implement it directly through prompting. After every evidence-collection action, the agent is instructed to review the updated evidence and determine whether another search is necessary before selecting its next action.

\subsubsection{Memory}

After reasoning on a labeled training node terminates, the agent stores the resulting trajectory in its private memory. Since the ground-truth label is available during this stage, we assign a binary reward according to whether the prediction is correct,
\begin{equation}
r_v = \mathbb{1}[\hat{y}_v = y_v].
\label{eq:reward}
\end{equation}
The complete reasoning process is recorded as
\begin{equation}
\tau_v =
\big(
z_v^{0},
s_v^{0}, a_v^{0}, \ldots,
s_v^{T_v}, a_v^{T_v},
\hat{y}_v, y_v, r_v
\big),
\label{eq:traj}
\end{equation}
where $T_v \leq T-1$ denotes the step at which $a_{\mathrm{pred}}$ is selected. We explicitly store the initial structural signature $z_v^{0}$ with each trajectory, which will later be used to retrieve past cases with similar structural patterns during collaborative reasoning.

After all labeled nodes in community $c_i$ have been processed, agent $A_i$ obtains its trajectory memory
\begin{equation}
\mathcal{M}^{\mathrm{traj}}_i
=
\left\{
\tau_v
\mid
v \in V_L \cap c_i
\right\},
\label{eq:memory}
\end{equation}
where $V_L$ denotes the set of labeled training nodes. The private memory of agent $A_i$ is written as
$\mathcal{M}_i=(\mathcal{M}^{\mathrm{traj}}_i,\mathcal{M}^{\mathrm{exp}}_i)$.
This stage only populates $\mathcal{M}^{\mathrm{traj}}_i$, while
$\mathcal{M}^{\mathrm{exp}}_i$ stores the learned experiences produced by the experience learning stage in Section~\ref{sec:stage4}. Since agents reason only over their assigned communities and do not communicate in this stage, their specialization processes can be performed in parallel. Algorithm~\ref{alg:independent} summarizes the complete agent specialization process.

\begin{algorithm}[t]
\caption{Agent Specialization}
\label{alg:independent}
\begin{algorithmic}[1]
\Require communities $\{c_i\}_{i=1}^{M}$; agents $\{A_i\}_{i=1}^{M}$; labeled node set $V_L$; step budget $T$
\Ensure trajectory memories $\{\mathcal{M}^{\mathrm{traj}}_i\}_{i=1}^{M}$

\For{each agent $A_i$, $i=1,\ldots,M$, \textbf{in parallel}}
    \State $\mathcal{M}^{\mathrm{traj}}_i \gets \emptyset$
    \State $\mathcal{M}^{\mathrm{exp}}_i \gets \emptyset$

    \For{each node $v \in V_L \cap c_i$}
        \State initialize sampled node set $O_v^{0}$ and compute structural signature $z_v^{0}$
        \State $\mathcal{X}_v^{0,k} \gets \emptyset$
        \State $s_v^{0} \gets (v,\mathbf{x}_v,\mathrm{text}(z_v^{0}),\mathcal{X}_v^{0,k})$

        \For{$t=0,1,\ldots,T-1$}
            \State select $a_v^{t}=\pi_i(s_v^{t},\mathcal{M}_i)$ by Eq.~\eqref{eq:policy}
            \Comment{force $a_{\mathrm{pred}}$ when $t=T-1$}

            \If{$a_v^{t}=a_{\mathrm{pred}}$}
                \State output $\hat{y}_v$ by Eq.~\eqref{eq:predict}
                \State $T_v \gets t$
                \State \textbf{break}
            \EndIf

            \State execute $a_v^{t}$ to sample additional nodes by Definitions~\ref{def:rhop} and~\ref{def:sem}
            \State update $O_v^{t+1}$ with the newly sampled nodes
            \State update structural signature $z_v^{t+1}$ by Eq.~\eqref{eq:sigupdate}
            \State rank sampled textual evidence and retain $\mathcal{X}_v^{t+1,k}$
            \State update observation $s_v^{t+1}$ by Eq.~\eqref{eq:stateupdate}
            \State review the current evidence before deciding whether another search is needed
        \EndFor

        \State $r_v \gets \mathbb{1}[\hat{y}_v=y_v]$
        \State construct $\tau_v$ by Eq.~\eqref{eq:traj}
        \State $\mathcal{M}^{\mathrm{traj}}_i \gets
        \mathcal{M}^{\mathrm{traj}}_i \cup \{\tau_v\}$
    \EndFor
\EndFor

\State \Return $\{\mathcal{M}^{\mathrm{traj}}_i\}_{i=1}^{M}$
\end{algorithmic}
\end{algorithm}

\subsection{Collaborative Reasoning}
\label{sec:stage3}

After agent specialization, each agent has developed its own reasoning experience within the assigned community and stored historical trajectories in its trajectory memory. We next describe how these specialized agents collaborate on cases that are difficult for a single agent.

\noindent\textbf{Collaboration Trigger.}
After agent $A_i$ produces a prediction for node $v$, we estimate how reliable the prediction is from its performance on structurally similar cases. Specifically, we retrieve the $m$ trajectories in $\mathcal{M}^{\mathrm{traj}}_i$ whose stored structural signatures are most similar to the initial signature $z_v^{0}$. The confidence of agent $A_i$ is defined as their similarity-weighted success rate,
\begin{equation}
\mathrm{conf}_i(v)
=
\frac{
\sum_{\tau \in \mathcal{Q}_i^{m}(v)}
\mathrm{sim}(z_v^{0},z_{\tau})\,r_{\tau}
}{
\sum_{\tau \in \mathcal{Q}_i^{m}(v)}
\mathrm{sim}(z_v^{0},z_{\tau})
},
\label{eq:conf}
\end{equation}
where $\mathcal{Q}_i^{m}(v) \subseteq \mathcal{M}^{\mathrm{traj}}_i$ denotes the retrieved trajectories, $z_{\tau}$ is the initial structural signature stored with trajectory $\tau$, and $r_{\tau}$ is its reward. We use cosine similarity for $\mathrm{sim}(\cdot,\cdot)$. A high value indicates that agent $A_i$ has frequently succeeded on nodes with similar structural signatures. Collaboration is triggered when $\mathrm{conf}_i(v)<\delta$, where $\delta$ is a predefined threshold.

\noindent\textbf{Collaborator Selection.}
When collaboration is triggered, we compute the same confidence score for every other agent using its own trajectory memory. Agents that have performed well on structurally similar cases are preferred as collaborators. Specifically, we select
\begin{equation}
\mathcal{K}(v)
=
\operatorname*{arg\,top\text{-}K}_{j\neq i}
\mathrm{conf}_j(v),
\label{eq:collaborator}
\end{equation}
where $\mathrm{conf}_j(v)$ is computed from $\mathcal{M}^{\mathrm{traj}}_j$ using Eq.~\eqref{eq:conf}. In this way, collaborator selection directly reuses the region-specific experience accumulated during agent specialization and requires no additional training.

\noindent\textbf{Collaborative Prediction.}
The owning agent $A_i$ and the selected agents in $\mathcal{K}(v)$ then perform debate-style collaborative reasoning \cite{du2023improving}. In the first round, each collaborator receives the target node $(v,\mathbf{x}_v)$ together with its initial structural signature $z_v^{0}$ and independently runs the reasoning process described in Section~\ref{sec:stage2}. Each agent instead collects and represents its own structural and semantic evidence using its private policy and memory, thereby preserving the specialization learned from its assigned community.

After the first round, each participant outputs a prediction together with a short rationale. In the following rounds, the participants read the previous-round predictions and rationales and independently reconsider their own decisions without collecting additional graph evidence. Let $\hat{y}_v^{(j,\ell)}$ denote the prediction of agent $A_j$ at debate round $\ell$. After $D$ rounds, the final prediction is obtained by majority vote,
\begin{equation}
\hat{y}_v
=
\arg\max_{y\in\mathcal{Y}}
\left|
\left\{
j\in\{i\}\cup\mathcal{K}(v)
:
\hat{y}_v^{(j,D)}=y
\right\}
\right|.
\label{eq:vote}
\end{equation}
If multiple labels receive the same number of votes, we choose the prediction from the tied participant with the highest confidence score. Algorithm~\ref{alg:reasoning} summarizes the complete collaborative reasoning process. Fig.~\ref{fig:prompt_debate} shows the prompt template used in the debate rounds.

\begin{figure}[t]
\centering
\fbox{\parbox{0.96\columnwidth}{\footnotesize\ttfamily\raggedright
You are agent \{id\}, one of several agents classifying the same \{item\} in a \{graph kind\}. In the previous round every agent reasoned independently. First review the other agents' opinions below, then give your own updated prediction independently. You may keep or revise your previous answer; do not follow the majority blindly, follow the evidence.\\[4pt]
Target \{item\}: "\{node text\}"\\[4pt]
Structural signature of the target node:\\
\{signature dimensions and observed label counts, as in the reasoning prompt\}\\[4pt]
Previous-round opinions:\\
\{for each participant: agent id; predicted label; brief rationale\}\\[4pt]
Candidate categories: [\{label list\}]\\[4pt]
Respond strictly in JSON: \{"thought": "...", "label": "one candidate category"\}
}}
\caption{Prompt template for debate rounds after the first round. Each participant reads the previous-round opinions and independently updates its prediction.}
\label{fig:prompt_debate}
\end{figure}

\noindent\textbf{Communication Cost.}
A direct transfer of conventional multi-agent collaboration to graphs would require agents to exchange the sampled graph evidence in textual form. For an $h$-hop neighborhood, this context grows with the number of sampled nodes and their textual attributes, which can lead to substantial token costs. In \textsc{MAAGL}, structural evidence is communicated through the fixed-dimensional signature $z_v^{0}$, while sampled semantic evidence remains local to each agent. During debate, agents exchange only their predictions and length-bounded rationales. Therefore, the communication cost does not grow with the size of the sampled graph neighborhood.

\begin{algorithm}[t]
\caption{Collaborative Reasoning}
\label{alg:reasoning}
\begin{algorithmic}[1]
\Require target node $v$; owning agent $A_i$; agents $\{A_j\}_{j=1}^{M}$; trajectory memories $\{\mathcal{M}^{\mathrm{traj}}_j\}_{j=1}^{M}$; threshold $\delta$; collaborator number $K$; reasoning budget $T$; debate rounds $D$
\Ensure final prediction $\hat{y}_v$

\State initialize $z_v^{0}$ and observation $s_v^{0}$
\State $A_i$ independently reasons on $v$ using Algorithm~\ref{alg:independent} and outputs $\hat{y}_v^{(i,1)}$
\State compute $\mathrm{conf}_i(v)$ from $\mathcal{M}^{\mathrm{traj}}_i$ by Eq.~\eqref{eq:conf}

\If{$\mathrm{conf}_i(v) \geq \delta$}
    \State \Return $\hat{y}_v^{(i,1)}$
\EndIf

\For{each agent $A_j$, $j\neq i$}
    \State compute $\mathrm{conf}_j(v)$ from $\mathcal{M}^{\mathrm{traj}}_j$ by Eq.~\eqref{eq:conf}
\EndFor

\State select collaborators $\mathcal{K}(v)$ by Eq.~\eqref{eq:collaborator}

\For{each $A_j \in \mathcal{K}(v)$ \textbf{in parallel}}
    \State initialize $s_{v,j}^{0}$ using $(v,\mathbf{x}_v,z_v^{0})$
    \State $A_j$ independently collects and represents its own graph evidence
    \State output $\hat{y}_v^{(j,1)}$ with a short rationale
\EndFor

\For{$\ell=2,\ldots,D$}
    \For{each participant $A_j$, $j\in\{i\}\cup\mathcal{K}(v)$, \textbf{in parallel}}
        \State read the round-$(\ell-1)$ predictions and rationales
        \State output revised prediction $\hat{y}_v^{(j,\ell)}$ with a short rationale
    \EndFor
\EndFor

\State obtain $\hat{y}_v$ by majority vote using Eq.~\eqref{eq:vote}
\State \Return $\hat{y}_v$
\end{algorithmic}
\end{algorithm}

\subsection{Experience Learning}
\label{sec:stage4}

After agent specialization and collaborative reasoning, each agent has accumulated trajectories that record the sampled evidence, reasoning actions, predictions, and corresponding rewards. These trajectories provide more than individual successful or failed cases: they can reveal recurring relations between graph characteristics and effective reasoning strategies. We therefore further extract reusable experience from the trajectory memories and use it to guide subsequent reasoning.

Existing experience-learning methods for LLM agents typically store experience as free-form natural language. For graph reasoning, however, such descriptions can be ambiguous. For example, an experience such as ``search for semantic evidence when nearby labels disagree'' does not specify how disagreement is measured or when the recommendation should be applied. Since \textsc{MAAGL} already represents structural evidence through explicitly defined signature metrics, we instead express each experience as a quantitative condition together with a reasoning recommendation.

\noindent\textbf{Experience Generation.}
Each agent first learns experience independently from its own trajectory memory. Following the contrastive learning principle of ExpeL \cite{zhao2024expel}, agent $A_i$ is provided with successful and failed trajectories from $\mathcal{M}^{\mathrm{traj}}_i$, together with the names and descriptions of the structural signature metrics. The agent compares these trajectories and identifies structural conditions that are associated with different reasoning outcomes. Each generated experience is represented as
\begin{equation}
\epsilon = (g,\theta,\rho)
\label{eq:experience}
\end{equation}
where $g$ is one of the signature metrics, $\theta$ specifies a threshold on that metric, and $\rho$ is a recommended reasoning strategy when the condition $g>\theta$ holds. The generated experiences are stored in the private experience memory $\mathcal{M}^{\mathrm{exp}}_i$ of agent $A_i$. Together with the trajectory memory, the complete private memory is therefore $\mathcal{M}_i=(\mathcal{M}^{\mathrm{traj}}_i,\mathcal{M}^{\mathrm{exp}}_i)$. The trajectory memory provides concrete past cases, whereas the experience memory provides reusable reasoning guidance distilled from these cases.

\noindent\textbf{Cross-Agent Experience Validation.}
An experience learned from one community may reflect only its local graph patterns and may not generalize to other regions. We therefore validate each generated experience using the trajectories of the other agents before making it globally available. Importantly, this validation operates directly on stored trajectories and does not require additional LLM calls.

Consider an experience $\epsilon=(g,\theta,\rho)$ generated by agent $A_i$. For another agent $A_j$, let $\mathcal{T}_j^{+}(\epsilon)$ denote the trajectories whose initial structural signatures satisfy $g(z_\tau)>\theta$ and which take the recommended action $\rho$ at some step, and let $\mathcal{T}_j^{-}(\epsilon)$ contain trajectories satisfying the same structural condition but never taking $\rho$. We measure the effectiveness of the recommendation on agent $A_j$ as
\begin{equation}
\Delta_j(\epsilon)
=
\hat{p}_j(r=1\mid \mathcal{T}_j^{+}(\epsilon))
-
\hat{p}_j(r=1\mid \mathcal{T}_j^{-}(\epsilon))
\label{eq:separation}
\end{equation}
where $\hat{p}_j(\cdot)$ denotes the empirical success rate over the corresponding trajectories. A positive $\Delta_j(\epsilon)$ indicates that, for nodes satisfying the structural condition, trajectories following $\rho$ achieve a higher success rate.

We admit an experience to the shared experience set $\mathcal{S}$ when it is supported by sufficiently many other agents,
\begin{equation}
\left|
\left\{
j\neq i :
\Delta_j(\epsilon)\geq\gamma
\right\}
\right|
\geq\kappa
\label{eq:admit}
\end{equation}
where $\gamma$ controls the minimum improvement required from the recommended strategy and $\kappa$ specifies the minimum number of supporting agents. Experiences that pass this validation are added to $\mathcal{S}$ and can be used by all agents in subsequent reasoning, while agent-specific experiences remain in their corresponding private experience memories. Example~\ref{ex:validation} illustrates the validation process, and Algorithm~\ref{alg:learning} summarizes the complete experience learning process.

\medskip
\noindent\fbox{\parbox{0.96\columnwidth}{\small
\refstepcounter{example}\label{ex:validation}%
\textbf{\textit{Example \theexample\ (cross-agent validation).}}
Suppose agent $A_1$ generates the experience
$\epsilon=(\text{label entropy},\,0.8,\,\text{semantic search})$.
For the trajectories of agent $A_2$ with label entropy above $0.8$, those that performed semantic search achieve a higher success rate than those that did not, giving $\Delta_2(\epsilon)=0.27$. Similarly, $\Delta_3(\epsilon)=0.23$. If $\gamma=0.2$ and $\kappa=2$, the experience is supported by two other agents and is therefore added to the shared experience set $\mathcal{S}$.
}}
\medskip

\begin{algorithm}[htbp]
\caption{Experience Learning}
\label{alg:learning}
\begin{algorithmic}[1]
\Require trajectory memories $\{\mathcal{M}^{\mathrm{traj}}_i\}_{i=1}^{M}$; experience memories $\{\mathcal{M}^{\mathrm{exp}}_i\}_{i=1}^{M}$; signature metrics $\{g_1,\ldots,g_d\}$; shared experience set $\mathcal{S}$; thresholds $\gamma$ and $\kappa$
\Ensure updated $\{\mathcal{M}^{\mathrm{exp}}_i\}_{i=1}^{M}$ and $\mathcal{S}$

\For{each agent $A_i$, $i=1,\ldots,M$, \textbf{in parallel}}
    \State sample successful and failed trajectories from $\mathcal{M}^{\mathrm{traj}}_i$
    \State generate candidate experiences $I_i=\{\epsilon=(g,\theta,\rho)\}$ from the sampled trajectories
    \State $\mathcal{M}^{\mathrm{exp}}_i \gets \mathcal{M}^{\mathrm{exp}}_i \cup I_i$
\EndFor

\For{each experience $\epsilon\in I_i$ generated by agent $A_i$}
    \For{each agent $A_j$, $j\neq i$}
        \State compute $\Delta_j(\epsilon)$ from $\mathcal{M}^{\mathrm{traj}}_j$ by Eq.~\eqref{eq:separation}
    \EndFor
    \If{Eq.~\eqref{eq:admit} holds}
        \State $\mathcal{S}\gets\mathcal{S}\cup\{\epsilon\}$
    \EndIf
\EndFor

\State \Return $\{\mathcal{M}^{\mathrm{exp}}_i\}_{i=1}^{M}$ and $\mathcal{S}$
\end{algorithmic}
\end{algorithm}

\noindent\textbf{Computation Cost.}
The structural signature introduces only limited additional computation. The four static dimensions are computed once before reasoning and cached for subsequent episodes. During reasoning, only the two dynamic dimensions need to be updated as new labeled nodes are sampled. Since these dimensions depend on label counts, they can be updated incrementally from the newly collected evidence without recomputing the entire sampled neighborhood. Confidence estimation in Eq.~\eqref{eq:conf} requires comparing the current initial signature with stored trajectory signatures, with a direct cost of $O(|\mathcal{M}^{\mathrm{traj}}_i|d)$ for agent $A_i$. Cross-agent collaborator selection and experience validation reuse the same cached signatures and stored trajectory outcomes and therefore require no additional graph sampling or LLM calls.

\section{Experiments}
We conduct extensive experiments to answer the following research questions (RQs):
\begin{itemize}
    \item \textbf{RQ1:} How does \textsc{MAAGL} compare with existing state-of-the-art AGL methods?
    \item \textbf{RQ2:} How does each core component contribute to the overall performance?
    \item \textbf{RQ3:} How sensitive is \textsc{MAAGL} to key hyperparameters?
    \item \textbf{RQ4:} How efficient is \textsc{MAAGL} in terms of reasoning cost?
\end{itemize}

\subsection{Datasets}
\begin{table}[t]
\centering
\caption{Statistics of the datasets.}
\label{tab:datasets}
\resizebox{\columnwidth}{!}{%
\begin{tabular}{llrrrr}
\toprule
Domain & Dataset & \#Nodes & \#Edges & \#Classes & Setting \\
\midrule
\multirow{2}{*}{Academic} & OGB-Arxiv & 169,343 & 1,166,243 & 40 & In-Domain \\
                          & Cora-full & 19,793 & 126,842 & 70 & Zero-shot \\
\midrule
\multirow{2}{*}{E-commerce} & OGB-Products & 54,025 & 74,420 & 47 & In-Domain \\
                            & Amazon-Computers & 87,229 & 1,256,548 & 10 & Zero-shot \\
\bottomrule
\end{tabular}%
}
\end{table}

We evaluate \textsc{MAAGL} on four TAGs spanning two domains. For each domain, we use one dataset for \emph{in-domain} training and evaluation, and a second, disjoint dataset from the \emph{same} domain to assess \emph{zero-shot transfer}. In the academic citation domain, nodes represent papers, edges represent citation relations, and the node text is the title and abstract of the paper; we train on OGB-Arxiv~\cite{hu2020open} and transfer to Cora-full~\cite{bojchevski2017deep,shchur1811pitfalls}. In the e-commerce domain, nodes represent products, edges connect products that are frequently bought together, and the node text is the title and description of the product in OGB-Products and a user review of the product in Amazon-Computers; we train on the OGB-Products subset of TAPE~\cite{hu2020open,he2024harnessing} and transfer to Amazon-Computers~\cite{yan2023comprehensive}. The statistics of all datasets are summarized in Table~\ref{tab:datasets}.

\begin{table*}[t]
\centering
\caption{Overall node classification performance (Accuracy and Macro-F1) under the in-domain and zero-shot transfer settings.}
\label{tab:main}
\resizebox{0.85\textwidth}{!}{%
\begin{tabular}{c l cc cc cc cc}
\toprule
\multicolumn{2}{l}{Settings} & \multicolumn{4}{c}{In-Domain} & \multicolumn{4}{c}{Zero-shot Transfer} \\
\cmidrule(lr){3-6} \cmidrule(lr){7-10}
\multicolumn{2}{l}{Datasets} & \multicolumn{2}{c}{OGB-Arxiv} & \multicolumn{2}{c}{OGB-Products} & \multicolumn{2}{c}{Cora-full} & \multicolumn{2}{c}{Amazon-Comp.} \\
\cmidrule(lr){3-4} \cmidrule(lr){5-6} \cmidrule(lr){7-8} \cmidrule(lr){9-10}
Category & Methods & Acc & Ma-F1 & Acc & Ma-F1 & Acc & Ma-F1 & Acc & Ma-F1 \\
\midrule
\multirow{3}{*}{\shortstack{\textit{GNN}\\\textit{Methods}}}
 & GCN \cite{kipf2016semi}          & 0.6460 & 0.3521 & 0.6710 & 0.4014 & -- & -- & -- & -- \\
 & GraphSAGE \cite{hamilton2017inductive} & 0.6210 & 0.3766 & 0.6610 & 0.4151 & -- & -- & -- & -- \\
 & GraphGPS \cite{rampavsek2022recipe} & 0.5530 & 0.2044 & 0.5170 & 0.2635 & -- & -- & -- & -- \\
\midrule
\multirow{4}{*}{\shortstack{\textit{Single-Agent}\\\textit{Methods}}}
 & Graph-CoT \cite{jin2024graph}   & 0.6240 & 0.4165 & 0.3690 & 0.2902 & 0.5780 & 0.5399 & 0.8340 & 0.8657 \\
 & ReaGAN \cite{guo2025reagan}      & 0.6390 & 0.4764 & 0.7210 & 0.5074 & 0.6447 & 0.5974 & 0.7830 & 0.8306 \\
 & AgentGL \cite{sun2026agentgl}    & 0.6550 & \textbf{0.4851} & 0.6860 & 0.5011 & 0.6320 & 0.5850 & 0.7100 & 0.6495 \\
 & GraphReAct \cite{yu2026graphreact} & 0.6320 & 0.4389 & 0.6650 & 0.4821 & 0.6260 & 0.5624 & 0.8110 & 0.8179 \\
\midrule
\multirow{2}{*}{\shortstack{\textit{Orchestration-based}\\\textit{Methods}}}
 & GraphTeam \cite{li2024graphteam}    & 0.6200 & 0.4487 & 0.5990 & 0.3909 & 0.5950 & 0.5457 & 0.6180 & 0.5500 \\
 & GraphAgent \cite{yang2025graphagent} & 0.6531 & 0.4652 & 0.6730 & 0.4768 & 0.6170 & 0.5762 & 0.7110 & 0.6856 \\
\midrule
\textit{Ours} & \textbf{\textsc{MAAGL}}   & \textbf{0.6690} & 0.4832 & \textbf{0.7650} & \textbf{0.5588} & \textbf{0.6590} & \textbf{0.6035} & \textbf{0.8670} & \textbf{0.8902} \\
\bottomrule
\end{tabular}%
}
\end{table*}

\subsection{Baselines}
We compare \textsc{MAAGL} against three categories of baselines:
\begin{itemize}
    \item \textbf{GNN methods} learn node representations by message passing or graph transformers and are trained on each dataset separately.
    \begin{itemize}
        \item \textbf{GCN}~\cite{kipf2016semi} aggregates the normalized features of the 1-hop neighbors in every layer.
        \item \textbf{GraphSAGE}~\cite{hamilton2017inductive} aggregates sampled neighbors with a learnable function.
        \item \textbf{GraphGPS}~\cite{rampavsek2022recipe} combines local message passing with global attention in a modular graph transformer.
    \end{itemize}

    \item \textbf{Single-agent methods} use a single LLM-powered agent to adaptively collect graph evidence and reason.
    \begin{itemize}
        \item \textbf{Graph-CoT}~\cite{jin2024graph} lets the LLM iteratively invoke graph functions to collect graph information.
        \item \textbf{ReaGAN}~\cite{guo2025reagan} equips an agent with neighbor-expansion and semantic retrieval tools, and predicts using the retrieved evidence.
        \item \textbf{AgentGL}~\cite{sun2026agentgl} structures graph reasoning into thought-action-observation trajectories and optimizes the agent policy through reinforcement learning.
        \item \textbf{GraphReAct}~\cite{yu2026graphreact} performs multi-step reasoning and acting for graph inference.
    \end{itemize}

    \item \textbf{Orchestration-based methods} coordinate multiple role-specific agents through a predefined workflow for graph reasoning.
    \begin{itemize}
        \item \textbf{GraphTeam}~\cite{li2024graphteam} coordinates role-specialized agents to collaboratively solve graph reasoning tasks.
        \item \textbf{GraphAgent}~\cite{yang2025graphagent} organizes multiple role-specific agents into a predefined workflow for graph reasoning.
    \end{itemize}
\end{itemize}

\subsection{Experimental Setup}
Following AgentGL~\cite{sun2026agentgl}, on the two in-domain training datasets (OGB-Arxiv and OGB-Products) we sample 3,000 training nodes each for optimization, and for each dataset we sample 1,000 nodes from the original test split for evaluation. The supervised baselines (GCN, GraphSAGE, and GraphGPS) are trained on the same 3,000 training nodes, using the node features shipped with each dataset, with model selection on the official validation split; since their label space is tied to the training graph, they have no zero-shot transfer results. For the zero-shot transfer datasets (Cora-full and Amazon-Computers), no training is performed on the target graph and we directly evaluate on 1,000 sampled test nodes: \textsc{MAAGL} reuses the shared experiences and the trajectory memories learned on the in-domain dataset of the same domain, which transfer unchanged because they contain structural signatures, reasoning actions, and correctness outcomes but no label names. \textsc{MAAGL} adopts GPT-4o-mini as the backbone of every agent and is fully training-free: it relies solely on the accumulated trajectories and the learned experiences, without updating any model parameter. Unless stated otherwise, \textsc{MAAGL} partitions the graph with Leiden into $M=10$ communities and uses $K=2$ collaborators, $D=2$ debate rounds, $R=2$ learning rounds, and a confidence threshold $\delta$ set to the 25th percentile of the training confidences.

\subsection{Overall Performance (RQ1)}
We first evaluate the overall performance of \textsc{MAAGL} against all baselines on node classification, under both the in-domain and the zero-shot transfer settings. The results are reported in Table~\ref{tab:main}, where baselines are grouped by category and our method is listed at the bottom. We summarize the key observations below.
\begin{itemize}
    \item \textsc{MAAGL} achieves the best Accuracy on all four datasets. For Macro-F1, it achieves the best results on OGB-Products and Amazon-Comp., while remaining competitive with the best baselines on OGB-Arxiv and Cora-full. These results demonstrate the overall effectiveness of \textsc{MAAGL}. A possible reason is that region-specific specialization allows each agent to learn from structurally and semantically similar cases, while selective collaboration further introduces complementary reasoning when the owning agent is less reliable.

    \item Agent-based methods generally outperform conventional GNN methods, particularly in terms of Macro-F1. This advantage is also reflected in the zero-shot transfer setting, where trained GNN models cannot be directly transferred to unseen graphs, whereas agent-based methods remain applicable without retraining. This observation supports the motivation of agentic graph learning, where LLM agents can adaptively collect graph evidence and exploit textual semantics for different instances, providing stronger generalization across graphs.

    \item Existing orchestration-based methods do not consistently outperform single-agent methods. For example, GraphAgent performs competitively on OGB-Arxiv but falls behind strong single-agent methods on the other datasets. This observation is consistent with our motivation that simply organizing multiple agents does not fundamentally address the limitation of applying a shared reasoning policy across different graph regions. In contrast, \textsc{MAAGL} consistently improves upon both categories by allowing agents to specialize in different regions and collaborate selectively when complementary experience is needed.
\end{itemize}

\subsection{Ablation Study (RQ2)}
\begin{figure}[t]
\centering
\includegraphics[width=\columnwidth]{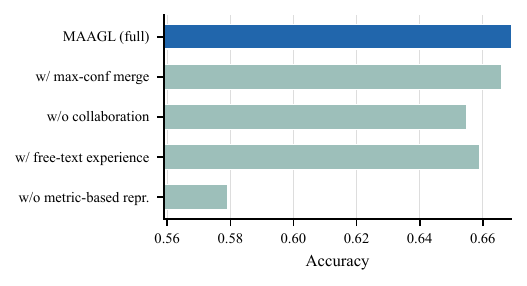}
\caption{Ablation study on node classification on OGB-Arxiv.}
\label{fig:ablation}
\end{figure}

To understand the contribution of each core component, we compare \textsc{MAAGL} with four ablated variants on OGB-Arxiv, each removing or replacing exactly one design choice: (i) \emph{w/o structural signature}, which verbalizes the retrieved neighbors as raw text in the ReaGAN style and removes the structural signature from the observation, while the numeric signature is still used for the collaboration trigger and collaborator selection so that only the representation effect is isolated; (ii) \emph{w/ free-text experience}, which replaces the structured experiences with free-text experiences in natural language kept in each agent's private experience memory, without cross-agent validation; (iii) \emph{w/o collaboration}, where the owning agent always answers alone; and (iv) \emph{w/ max-confidence merge}, which keeps collaboration but replaces the debate and the vote with the single prediction of the most confident participant. Figure~\ref{fig:ablation} reports Accuracy on node classification.

\begin{itemize}
    \item Removing the structural signature causes the largest performance degradation. This confirms the importance of separating structural and semantic evidence instead of directly verbalizing sampled neighborhoods, supporting our motivation that raw textual serialization is unsuitable for graph reasoning.

    \item Replacing the structured experience with free-text experience consistently weakens performance. This suggests that expressing experience through explicit structural conditions and recommendations provides more reliable guidance than unconstrained natural-language lessons.

    \item Removing collaboration also degrades performance, showing that independently specialized agents are not sufficient for all instances. Selective collaboration allows agents to incorporate complementary experience from other regions when their own reasoning is less reliable.

    \item Replacing debate-based collaboration with max-confidence merging leads to a smaller performance drop. This indicates that simply selecting the most confident prediction cannot fully exploit the complementary reasoning of multiple agents, while iterative exchange and refinement provides additional benefit.
\end{itemize}

\subsection{Hyperparameter Sensitivity (RQ3)}
\label{sec:rq3}
\begin{figure}[t]
    \centering
    \subfloat[Community detection algorithm]{\includegraphics[width=0.49\columnwidth]{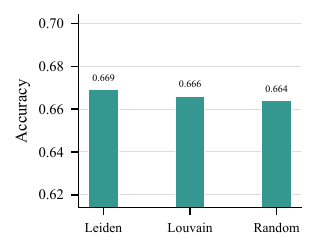}}
    \hfil
    \subfloat[Number of communities]{\includegraphics[width=0.49\columnwidth]{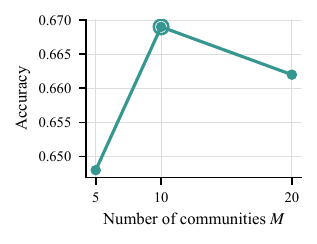}}
    \\
    \subfloat[Number of debate rounds]{\includegraphics[width=0.49\columnwidth]{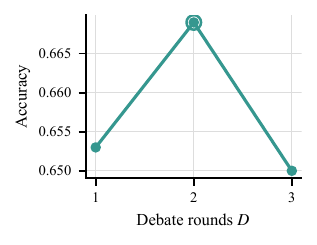}}
    \hfil
    \subfloat[Confidence threshold]{\includegraphics[width=0.49\columnwidth]{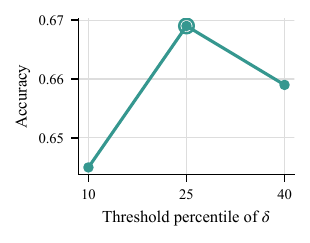}}
    \caption{Hyperparameter sensitivity of \textsc{MAAGL} on OGB-Arxiv.}
    \label{fig:hyper}
\end{figure}

\begin{figure}[t]
    \centering
    \includegraphics[width=1\columnwidth]{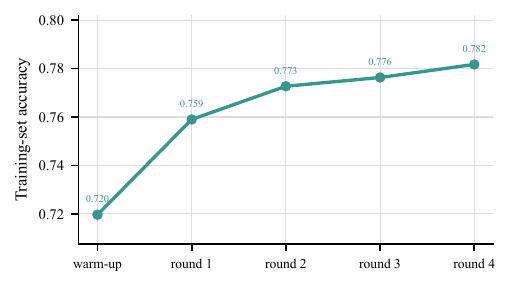}
    \caption{Accuracy on the learning set of OGB-Arxiv across learning rounds.}
    \label{fig:trial}
\end{figure}

We study the sensitivity of \textsc{MAAGL} to four key hyperparameters: (a) the choice of community detection algorithm including Leiden, Louvain, and random partition. (b) the number of communities $M \in \{5, 10, 20\}$). (c) the number of debate rounds ($D \in \{1, 2, 3\}$). (d) the confidence threshold that triggers collaboration ($\delta$ set to the 10th, 25th, or 40th percentile of the training confidences). Figure~\ref{fig:hyper} reports node classification Accuracy as each hyperparameter varies while the others are fixed at their default values. We also study the effect of the number of learning rounds (from the warm-up only up to $R=4$), in which collaborative reasoning and experience learning alternate. Figure~\ref{fig:trial} reports the accuracy on the training set of OGB-Arxiv as more learning rounds are performed.

\begin{itemize}
    \item \textsc{MAAGL} is relatively insensitive to the choice of community detection algorithm. Leiden, Louvain, and random partition all achieve comparable performance, while community-based partitioning provides a small advantage. This suggests that the benefit mainly comes from assigning different graph regions to independent agents rather than relying on a particular partitioning algorithm.
    \item The number of communities has a clear effect on performance. Too few communities limit agent specialization, while too many communities divide the graph into overly small regions and reduce the experience available to each agent. A moderate number of communities provides a better balance between specialization and sufficient local experience.
    \item A small number of debate rounds is sufficient for effective collaboration. Increasing the number of rounds initially improves performance by allowing agents to reconsider their predictions using complementary opinions, whereas further debate provides little additional benefit and may introduce unnecessary reconsideration.
    \item A moderate confidence threshold achieves the best performance. A low threshold triggers collaboration for too few cases, while a high threshold invokes collaboration even when the owning agent is already reliable. This supports selectively introducing collaboration for uncertain instances.
    \item Performance improves consistently over successive learning rounds and gradually approaches saturation. This indicates that agents can progressively accumulate useful experience from historical reasoning trajectories, while the diminishing improvement in later rounds suggests convergence of the learning process.
\end{itemize}

\begin{figure}[t]
    \centering
    \includegraphics[width=0.8\columnwidth]{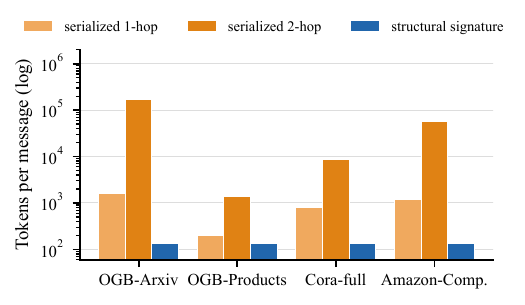}
    \caption{Tokens per cross-agent message on the four datasets.}
    \label{fig:efficiency}
\end{figure}

\begin{figure}[t]
    \centering
    \includegraphics[width=0.9\columnwidth]{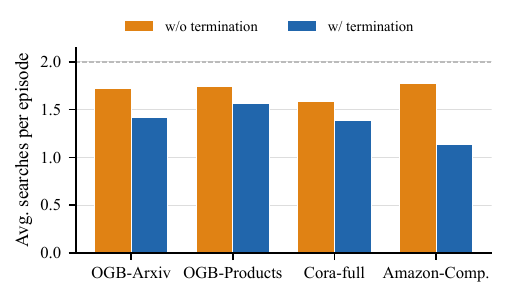}
    \caption{Average number of search actions per episode on each dataset, with and without the Adaptive Search Termination.}
    \label{fig:trigger}
\end{figure}

\subsection{Efficiency Analysis (RQ4)}
We analyze the two efficiency designs of \textsc{MAAGL} on all four datasets. Figure~\ref{fig:efficiency} compares the number of tokens one cross-agent message carries, when the message serializes the 1-hop or 2-hop neighborhood and when it carries only the structural signature, measured on the 1,000 test nodes of each dataset with 200 tokens per node text. Figure~\ref{fig:trigger} reports the average number of search actions per episode on each dataset, with and without adaptive search termination.
\begin{itemize}
    \item The structural signature substantially reduces the token cost of cross-agent communication across all datasets. Unlike serialized neighborhoods, whose size grows rapidly with the number of retrieved nodes and neighborhood depth, the signature remains compact and fixed in size. This confirms that separating structural evidence from raw textual content provides a more efficient representation for multi-agent communication.

    \item Adaptive search termination consistently reduces the number of search actions across all datasets. This indicates that many instances can be resolved without exhausting the full search budget, and allowing the agent to stop once sufficient evidence has been collected avoids unnecessary graph exploration and improves reasoning efficiency.
\end{itemize}

\section{Conclusion}
In this work, we revisited multi-agent collaboration for agentic graph learning and identified two key limitations in directly transferring existing collaborative reasoning methods to graphs. First, existing AGL methods generally rely on a shared reasoning policy across different graph regions, which can be suboptimal when these regions exhibit different structural and semantic patterns. Second, communicating graph evidence through natural-language serialization introduces arbitrary neighbor ordering and rapidly increasing token costs. To address these issues, we proposed \textsc{MAAGL}, which assigns independent agents to different graph communities for region-specific specialization and represents sampled structural and semantic evidence separately. Structural evidence is summarized by a permutation-invariant structural signature, while semantic evidence is filtered according to relevance. Historical trajectories are further used to estimate agent confidence, select complementary collaborators, and learn reusable reasoning experience. Extensive experiments on four benchmark datasets demonstrate that \textsc{MAAGL} consistently improves over existing AGL methods.

One limitation of the current framework is that the dimensions of the structural signature are manually specified based on commonly used graph statistics. An important direction for future work is therefore to automatically discover or select task-specific structural signature dimensions from graph data and reasoning trajectories.

\section*{Acknowledgment}
This work is supported by the Australian Research Council under the Discovery Project scheme (No.DP240101591).

% \vspace{10em}
\bibliographystyle{IEEEtran}
\bibliography{main}

\end{document}